\documentclass[runningheads]{llncs}
\usepackage{cite}
\usepackage{tabularx}
\usepackage{multirow}
\usepackage[T1]{fontenc}
\usepackage{booktabs}
\usepackage{amssymb}
\usepackage[dvipsnames,svgnames,x11names,table]{xcolor}
\usepackage{pifont}
\usepackage{tabularx}
\usepackage{floatrow}
\newfloatcommand{capbtabbox}{table}[][\FBwidth]
\usepackage{url}
\usepackage{amsmath}
\usepackage{bbm}
\usepackage{bm,upgreek}
\usepackage{verbatim}
\usepackage{enumitem}

\usepackage[belowskip=-15pt,aboveskip=0pt]{caption}

\usepackage{graphicx}
\makeatletter
\newcommand*\bigcdot{\mathpalette\bigcdot@{.5}}
\newcommand*\bigcdot@[2]{\mathbin{\vcenter{\hbox{\scalebox{#2}{$\m@th#1\bullet$}}}}}
\makeatother

\newcommand{\clip}{CLIP}
\newcommand{\mr}[1]{\mathit{#1}}

\newcommand{\loss}{\mathcal{L}}
\newcommand{\mypar}[1]{\vspace{0.35\baselineskip}\noindent\textbf{#1}\,}

\newcommand{\fImg}{\mathbf{I}}
\newcommand{\fTxt}{\mathbf{T}}

\newcommand{\fImgAtt}{\widetilde{\fImg}}

\newcommand{\xx}{\mathbf{x}}

\newcommand{\real}{\mathbb{R}}

\newcommand{\att}{a}

\newcommand{\vone}{\mathbbm{1}}

\newcommand{\PP}{\mathbf{P}}
\newcommand{\QQ}{\mathbf{Q}}
\newcommand{\SSS}{\mathbf{S}}

\newcommand{\data}{\mathcal{D}}

\newcommand{\tr}[1]{{#1^{\top}}}

\newsavebox\CBox
\def\textBF#1{\sbox\CBox{#1}\resizebox{\wd\CBox}{\ht\CBox}{\textbf{#1}}}

\begin{document}
\title{FACMIC: Federated Adaptative \clip{} Model for Medical Image Classification}
\titlerunning{Federated Adaptative \clip{} Model for Medical Image Classification}
%
\author{Yihang Wu\inst{1} \and
Christian Desrosiers\inst{2} \and
Ahmad Chaddad\inst{1,2}}
%
\institute{Laboratory for Artificial Intelligence for Personalised Medicine, School of Artificial Intelligence, Guilin University of Electronic Technology, China \and Laboratory for Imagery Vision and Artificial Intelligence, ÉTS Montreal, Canada \\
\email{ahmad8chaddad@gmail.com}}
\maketitle              

\begin{abstract}
Federated learning (FL) has emerged as a promising approach to medical image analysis that allows deep model training using decentralized data while ensuring data privacy. However, in the field of FL, communication cost plays a critical role in evaluating the performance of the model. Thus, transferring vision foundation models can be particularly challenging due to the significant resource costs involved. In this paper, we introduce a federated adaptive Contrastive Language Image Pretraining (\clip{}) model designed for classification tasks. We employ a light-weight and efficient feature attention module for \clip{} that selects suitable features for each client's data. Additionally, we propose a domain adaptation technique to reduce differences in data distribution between clients.
Experimental results on four publicly available datasets demonstrate the superior performance of FACMIC in dealing with real-world and multisource medical imaging data. Our codes are available at \url{https://github.com/AIPMLab/FACMIC}.

\end{abstract}
\section{Introduction}
The success of deep learning (DL) highly depends on the availability of large amounts of data for training. However, there has been a growing focus on data privacy and security in recent years, with some organizations implementing regulations and laws such as the EU General Data Protection Regulation (GDPR) \cite{voigt2017eu}. Collecting raw data is often impractical in this case, which poses a challenge to the feasibility of centralized DL approaches. Federated learning (FL) has emerged as a new distributed learning method to deal with this challenge and has been widely adopted in various applications \cite{chaddad2023federated}. FL enables model aggregation without directly accessing the raw user data from different clients. As pioneering work in FL, FedAVG efficiently combines distributed information through a simple but effective averaging algorithm \cite{li2019convergence}. This method ensures that raw data remain on the local client, preserving data privacy and security. 

The performance of FL models is typically influenced by two main factors: data distribution shifts and communication costs \cite{sattler2019robust}. Shifts in data distribution can negatively impact model prediction accuracy, particularly in the medical field, where variations in imaging equipment can introduce discrepancies. In most FL solutions, communication costs are proportional to the number of model parameters to transmit. This impedes the use of large Vision Language Models (VLMs) such as \clip{} that contain more than $10^8$ parameters (ViT-B/32) \cite{bommasani2021opportunities}.

Work on FL with foundation models has started recently. In \cite{guo2023promptfl}, the authors proposed sharing the prompts instead of models to reduce communication costs. An important limitation of this approach is that it does not address the issue of heterogeneity in the data distribution. In \cite{yu2023bridging}, a specialized model compression technique is introduced to reduce transmission costs. However, this technique still has substantial computational costs. Furthermore, the results in \cite{Huix_2024_WACV} show that foundation models like \clip{} or BLIP fail for medical image classification tasks (e.g., 52.7\% classification recall for \clip{} using the ISIC2019 dataset). This leads us to consider how federated foundation models can be enhanced in terms of effectiveness and efficiency. 
In this paper, we propose the Federated adaptive \clip{} model for the medical image classification (FACMIC) task. FACMIC adds a light-weight feature-attention module on top of \clip{} to help the model focus on useful features in the data for each client. It also introduces a domain adaptation (DA) strategy to minimize data discrepancies between clients. Our model can quickly converge and achieve improved performance through adaptation, which greatly reduces training time. The contributions of our work are summarized as follows.
\begin{itemize}[noitemsep,topsep=2pt]
    \item We propose a novel federated adaptive \clip{} model that combines feature attention with an adaptation technique to address both problems of communication costs and distribution discrepancy. Although past studies have applied FL in medical imaging, to our knowledge, we are the first to explore FL on VLMs in this context.
    \item We test the performance of our method in both real-world and simulated multi-source cases, and show its superior performance compared to state-of-the-art approaches for brain tumor and skin cancer classification tasks. 
\end{itemize}


\section{Material and methods}\label{S:2}

\subsection{Problem formulation}

In our federated learning setting, we have a set of $N$ clients $\{C_1, C_2, \cdots, C_N\}$, each client $C_i$ having a private data set $\data_i = \{(\xx_{i,j},\, y_{i,j})\}^{n_i}_{j=1}$. As in similar studies \cite{lu2023fedclip}, we assume that the data of separate clients have the same input and output space, but follow different distributions, i.e., $P(\data_{i'}) \neq P(\data_{i})$, $\forall i' \neq i$. Each dataset $\data_i$ consists of three non-overlapping parts, namely a training set $\data_i^{\mr{train}}$, a validation set $\data_i^{\mr{val}}$ and a test set $\data_i^{\mr{test}}$. Our goal is to train a robust global model $f_{\uptheta}(\cdot)$  while preserving data privacy and security. This model should provide good testing performance on the test data of every client, i.e.,
\begin{equation} \footnotesize
     \mathop {\min}\limits_f \ \frac{1}{N} \sum_{i=1}^N \frac{1}{n_i^{\mr{test}}} \sum_{j=1}^{n_i^{\mr{test}}} \ell(f_\uptheta(\xx^{\mr{test}}_{i,j}),\, y^{\mr{test}}_{i,j}),
\end{equation}
based on a given loss function $\ell$. For generalization, we assume that there exist $Q$ different clients $\{M_1,M_2,\cdots,M_Q\}$ with data $\data_i^M = \{(\xx_{i,j},\, y_{i,j})\}^{n_i}_{j=1}$, $n_i$ being the number of samples in each client. Our goal is for the model to achieve a good performance on clients that \emph{were excluded} from the local training stage, i.e.,
\begin{equation} \footnotesize
    \mathop {\min}\limits_f \ \frac{1}{M} \sum_{i=1}^M \frac{1}{m_i^{\mr{}}} \sum_{j=1}^{m_i^{\mr{}}} \ell(f_\uptheta(\xx^{\mr{}}_{i,j}),\, y^{\mr{}}_{i,j}).
\end{equation}

\subsection{Our federated adaptive \clip{} framework}

\begin{figure*}[t!]
    \centering
    \includegraphics[width = 0.89 \textwidth]{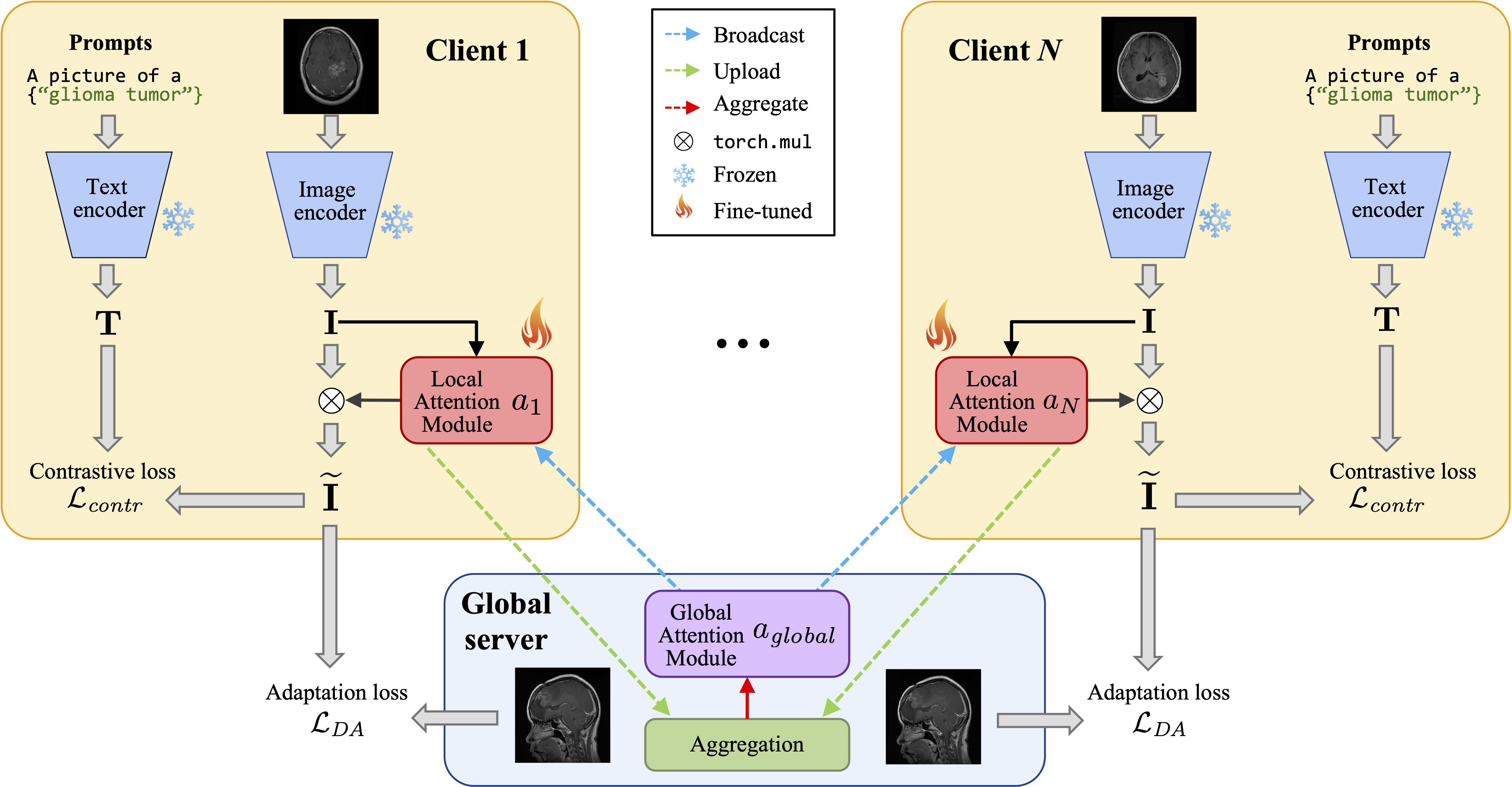}
    \caption{The proposed FACMIC framework. 
    Each client trains its model separately, optimizing only the parameters of its local attention module ($\att_i$) using contrastive and domain adaptation losses. After receiving the local client parameters, the server aggregates them into a global attention module ($\att_{\mr{global}}$) whose parameters are broadcasted back to clients.}
    \label{fig:framework}    
\end{figure*}

Our federated learning framework for \clip{}-based medical image classification is illustrated in Figure \ref{fig:framework}. This framework comprises three key components: a feature attention module for efficient FL, a feature adaptation strategy that addresses the problem of data distribution shifts across different clients, and a global aggregation strategy to combine the learned information from multiple clients. We present these components in what follows.

\mypar{Training the attention module.} We use a pretrained \clip{} model comprising an image encoder $e_I(\cdot)$ and a text encoder $e_T(\cdot)$, to extract features from the data for each client $C_i$. For a training example $\xx_j \in \data_{i}^{\mr{train}}$, we denote as $\fImg_j = e_I(\xx_j) \in \real^D$ the $D$-dimensional vector of image features. For text features, we use the standard prompt ``\texttt{a picture of a \{class\}}'' as input to the text encoder to obtain features $\fTxt_j = e_T(\xx_j) \in \real^D$. 

Pretrained foundation models hold the ability to extract a rich set of features, however, not all of those are suitable for learning a specific task. This is particularly true for detecting and classifying abnormal regions such as lesions in medical images, as these regions are absent in normal images and typically represent a small part of the image. To identify the regions of focus for locally-trained models, we introduce a client feature attention module, denoted as $\att_i(\cdot)$. This attention module takes as input image features $\fImg$ and returns an attention mask $\att_i(\fImg)\in [0,1]^D$. This mask is then used to generate masked images features $\fImgAtt = \att_i(\fImg) \otimes \fImg$, where $\otimes$ is the Hadamard (element-wise) product.

We measure the probability that an example $\xx_j$ belongs to a class $c$ using the cosine similarity between the image features of $\xx_j$ and the text features $\fTxt_c$ corresponding to the prompt of $c$:
\begin{equation}\label{eq:CLIP-classif}
p(\mathrm{Y}\!=\!c \, | \, \xx_j) \, =\,  \frac{\exp(s_{j,c}/\tau)}{\sum_{c'=1}^K \exp(s_{j,c'}/\tau)}, \ \ \text{with }
s_{j,c} = \frac{\langle\fImgAtt_j, \fTxt_c\rangle}{\|\fImgAtt_j\| \!\cdot\!\|\fTxt_c\|}
\end{equation}
where $\tau$ is the softmax temperature parameter. 

Keeping the image and text encoders frozen, we train the local adapter modules by minimizing a contrastive loss $\loss_{\mr{contr}}$ that pushes together the image and text features from the same training example and pulls apart non-matching ones. 
Following \cite{radford2021learning}, we compute the contrastive loss over batches of size $B$. Let $\SSS$ be the $B\!\times\!B$ matrix where $s_{j,j'}$ is the cosine similarity between image features $\fImgAtt_j$ and $\fTxt_{j'}$, as measured in Eq (\ref{eq:CLIP-classif}). We compute an image probability matrix $\PP = \mr{softmax}(\SSS/\tau) \in [0,1]^{B \times B}$ and a text probability matrix $\QQ = \mr{softmax}(\tr{\SSS}/\tau) \in [0,1]^{B \times B}$. The contrastive loss is then formulated as follows:
\begin{equation}
\loss_{\mr{contr}} \, = \, -\frac{1}{B}\sum_{j=1}^B \frac{1}{2}\Big(\log p_{j,j} \, + \, \log q_{j,j}\Big). 
\end{equation}

\mypar{Local feature adaptation.}
Discrepencies in the local data distribution of clients may affect the training of the global model via federated learning. To address this problem, we propose a client-specific feature adaptation strategy based on the Local Maximum Mean Discrepancy (LMMD) method \cite{zhu2020deep}. This domain adaptation method aligns the class-wise distribution statistics of the data from a source domain $\data_s$ and a target domain $\data_t$, by minimizing the following loss:
\begin{equation} \label{EQ:1}
    {\loss _{\mr{DA}}} = \frac{1}{K}\sum\limits_{c = 1}^K {\bigg\| {\sum\limits_{\xx_i^s \in {\data_s}} \!\!\! {\omega_{i,c}^{s}}\,\phi (\xx_i^s) \, - \!\!\!\sum\limits_{\xx_j^t \in {\data_t}} \!\!\!{\omega_{j,c}^{t}}\, \phi (\xx_j^t)} \bigg\|_\mathcal{H}^2}\!\!\!, \ \ \omega_{i,c}^{s|t}=\frac{\vone(y_{i}=c)}{\sum_{y_j\in\data^{s|t}} \vone(y_{j}=c)}. 
\end{equation}
Here, $\phi(\cdot)$ is a mapping function to a Hilbert space $\mathcal{H}$, while $\omega_{i,c}^s$, $\omega_{j,c}^t$ are weights that measure the membership of example $\xx_i^s$ and $\xx^t_j$ in class $c$.

In our setting, we suppose that the shift occurs only on the image features, thus $\xx\!=\!\fImg$ in Eq. (\ref{EQ:1}). Furthermore, since we cannot compute $\phi(\cdot)$ directly (as it maps features to a high-dimensional space), we use the kernel trick and reformulate the loss as
\begin{equation}
\begin{aligned}
{\loss_{\mr{DA}}}
= & \frac{1}{K}\sum\limits_{c = 1}^K {\Big[ {\sum\limits_{i = 1}^{{n_s}} {\sum\limits_{j = 1}^{{n_s}} {\omega_{i,c}^{s}} } \, \omega_{j,c}^{s} \,k(\fImg_i^s,\fImg_j^s)}}   \ + \ \sum\limits_{i = 1}^{{n_t}} {\sum\limits_{j = 1}^{{n_t}} {\omega_{i,c}^t} } \, \omega_{j,c}^{t}\, k(\fImg_i^t,\fImg_j^t) \\[-8pt]
& \qquad\qquad\qquad\qquad  { \ - \ 2\sum\limits_{i = 1}^{{n_s}} {\sum\limits_{j = 1}^{{n_t}} {\omega_{i,c}^{s}} } \,\omega_{j,c}^{t}\, k(\fImg_i^s,\fImg_j^t)} \Big] \\[-5pt]
\end{aligned}
\end{equation}
where $n_s, n_t$ are the number of samples in the source domain and target domain, respectively, and $k(\cdot,\cdot)=\langle\phi(\cdot),\phi(\cdot)\rangle$ is the kernel function. 

Although we defined our domain adaptation loss, we still need to specify the source and target domains used in this formulation. When performing a local adaption for client $C_i$, we set the source domain to be the data of this client, i.e., $\data_s=\data_i$. However, to preserve privacy, we cannot exchange data directly between clients, therefore we cannot use the data of other clients for the target domain. Instead, we use a global set of unlabeled images as the target domain data. This constraint can be easily satisfied since there are many publicly available datasets in medical imaging and no labels are needed for this reference data.

To compute the weights $\omega_{i,c}^{s|t}$ in Eq. (\ref{EQ:1}), we employ a pseudo-label strategy to estimate the class labels. Specifically, we obtain the class probabilities of a target image $\fImg^t_j$ using Eq. (\ref{eq:CLIP-classif}), and assign this image to the class with the highest probability. Our final loss to update the feature attention parameters of each client is the combination of the contrastive loss and domain adaptation loss,
\begin{equation}
\loss = \loss_{\mr{contr}} \, + \, \lambda \loss_{\mr{DA}},
\end{equation}
where $\lambda$ is a hyper-parameter controlling the trade-off between these two loss terms.

\mypar{Global aggregation.} The last part of our proposed FACMIC framework is the aggregation strategy to combine the parameters of different clients into a single global model. This strategy works as follows. In each round, each client $C_i$ uploads its attention module parameters $\uptheta^{\att}_i$ to the server. Thereafter, the server combines these parameters into a single vector $\uptheta^{\att}_{\mr{global}}$ using a weighted average
\begin{equation}\label{Eq:8}
    \uptheta^{\att}_{\mr{global}} = \sum\nolimits_{i = 1}^N \omega_i \cdot \uptheta^{\att}_i, \ \
    \omega_i = \frac{n_i^{\mr{train}}}{\sum\nolimits_{i' = 1}^N n_{i'}^{\mr{train}}}.
\end{equation}
Next, the server broadcasts the global attention module parameters back to each client. Since the attention module has only a small amount of parameters compared to the CLIP encoders, this strategy has very low communication and computational costs. 

\begin{table}[t!] \scriptsize
    \centering
    \caption{Samples of each client under non-iid conditions for BT and SC datasets.}
    \setlength{\tabcolsep}{5pt}
    \begin{tabular}{ccccccc}
    \toprule
     \multirow[b]{ 2}{*}{Clients}  &\multicolumn{2}{c}{$\alpha = 0.3$} &\multicolumn{2}{c}{ $\alpha = 0.6$}& \multicolumn{2}{c}{$\alpha = 0.9$}\\
     \cmidrule(l{4pt}r{4pt}){2-3}
     \cmidrule(l{4pt}r{4pt}){4-5}
     \cmidrule(l{4pt}r{4pt}){6-7}
     & BT & SC  & BT & SC & BT & SC  \\
     \midrule
     Client$1$ &2075&218&1355&387&1480&1156   \\
     Client$2$ &389&1374&908&487&423&335   \\
     Client$3$ &406&647&607&1365&967&748   \\
    Global &394&118 &394&118 &394&118   \\
     \bottomrule
    \end{tabular}
    \label{tab:my_label}
\end{table}

\vspace{-3pt}
\section{Experiments}\label{S:3}


\vspace{-3pt}
\subsection{Datasets}
\vspace{-3pt}

\mypar{Brain Tumor.}
We conduct experiments on a public MRI brain tumor classification dataset, denoted as BT \cite{BrainTumor2}. BT has four different classes, namely glioma tumor, meningioma tumor, no tumor, and pituitary tumor. The training set has 2,870 samples, while the testing set has 394 samples. 

\mypar{Skin Cancer.} We also use a public skin cancer (denoted as SC) dataset, obtained from The International Skin Imaging Collaboration (ISIC) \cite{Skin}. The data set contains the following diseases: actinic keratosis (AK), basal cell carcinoma (BCC), dermatofibroma (DF), melanoma (MEL), nevus (NV), pigmented benign keratosis (PBK), seborrheic keratosis (SK), squamous cell carcinoma (SCC) and vascular lesion (VL). The training set contains 2,239 samples, and the testing set 118 samples. For both the BT and SC dataset, we divide the training set into three clients, following both iid and non-iid conditions. 

\mypar{Real Multi-Source Skin Cancer.} We build this dataset (denoted as Real) from three sources, SC, HAM10000, and ISIC2019 \cite{tschandl2018ham10000,codella2018skin,combalia2019bcn20000}, where each source is treated as an individual client. Since ISIC2019 lacks a test set, we divided it into two parts, one for training and the other for testing, with a ratio of 8:2. We selected the common classes for this dataset: AK, BCC, DF, MEL, NV, PBK, and VL. The global testing set contains 6,233 samples, while Client $1$ (SC) has 1,971 samples, Client $2$ (ISIC2019) holds 19,766 samples and Client $3$ (HAM10000) possesses 8,512 samples. For the BT, SC and Real datasets, each client's data was then divided into three subsets: a training set, a validation set, and a testing set (8:1:1). Finally, we evaluated the global model using the global testing set.

\mypar{Data under iid condition.} For BT and SC, we randomly allocate the training set to each client with an equal number of samples. After this split, in BT, every client receives 956 samples, while in SC, every client holds 746 samples.

\mypar{Non-iid data.} We employ a Dirichlet distribution with concentration parameters $[\alpha, \alpha, \alpha]$, $\alpha \in \{0.3, 0.6, 0.9\}$, as a conjugate prior to generate non-iid data for the clients. For each class, we sample from this distribution and use the sampled values (one for each client) to divide the class examples among the clients. Table \ref{tab:my_label} shows the results after division for the BT and SC datasets.

\vspace{-3pt}
\subsection{Implementation details}
\vspace{-3pt}

The ViT-B/16 pre-trained model is adopted as the \clip{} backbone in our framework. Our light-weight attention module is composed of five layers: a first linear layer, a batch normalization layer, a LeakyReLU layer, a second linear layer, and a softmax activation function. During training, we keep the \clip{} encoders frozen and optimize only the attention module's parameters using Adam with beta parameters set to 0.9 and 0.98, a weight decay of 0.02, a fixed learning rate of 5$\times$$10^{-5}$, and a batch size of 32. 

For FL, we set the number of global training rounds to 100 for BT, 50 for SC and 50 for Real. For each round, we perform a single epoch of local training and aggregate the parameters of all clients. As image preprocessing, we resized the images to $224 \times 224$, and normalized their intensity using z-score normalization for both the training and testing phases \cite{fei2021z}. For the LMMD loss $\loss_{\mr{DA}}$, we adopt a Gaussian kernel with a bandwidth set to the median pairwise squared distances in the training data following \cite{zhu2020deep}, and use a weight of $\lambda=1$ for this loss term. The environment used for experiments is based on the Windows 11 operating system, and features an Intel 13900KF CPU with 128 GB of RAM and an RTX 4090 GPU.


\begin{table}[t!] \scriptsize
    \centering
    \caption{Global testing accuracy (ACC\%) and balanced accuracy (BACC\%) (highest) for BT, SC and Real dataset. The BACC metric is only adopted for Real dataset while the other results indicate ACC. The best results are marked in \textbf{bold}. Average indicates the mean value of ACC on all clients. Note: \ding{56} indicates no partition on data.}
    \setlength{\tabcolsep}{3pt}
    \begin{tabular}{l|cc|cc|cc|cc|cc|cc}
    \toprule
      \multirow[b]{2}{*}{Method} & \multicolumn{2}{c|}{$\alpha = 0.3$} & \multicolumn{2}{c|}{$\alpha = 0.6$} & \multicolumn{2}{c|}{$\alpha = 0.9$} & \multicolumn{2}{c|
}{$iid$} & \multicolumn{2}{c|}{Average} & \multicolumn{2}{c}{Real} \\
     \cmidrule(lr){2-13}
     & BT & SC & BT & SC & BT & SC & BT & SC & BT & SC &ACC &BACC \\
     \midrule
     Centralized & \ding{56} & \ding{56} & \ding{56} &\ding{56}&\ding{56}&\ding{56}&\ding{56}&\ding{56}  &70.30&\textbf{61.01} &60.81&47.80 \\
     \midrule
      FedAVG &66.24&48.30 &66.49&49.15 &64.21&54.23&68.52&49.15 &66.36&50.21 &60.24& 50.96 \\
     FedProx &66.24&48.30 &63.95&51.69 &65.23&49.15&70.05&50.84 &66.37 &49.99 &59.52&48.89 \\
     MOON &67.25&50.84 &63.96&50.00 &65.48&53.38&69.29&47.45 &66.49 &50.42 &59.95&50.90\\
      FedFocal &67.00&49.15 &65.48&49.15 &67.26&54.23&64.21&51.69  &65.99 &51.05 &59.97&49.80 \\
     Fed\clip{}&68.78&53.38 &66.24&55.93 & 67.26&54.23 &66.50&50.0 &67.19 &53.38 &58.43&51.17 \\
     \rowcolor{gray!15} Ours &\textBF{82.74}&\textBF{56.78} & \textBF{82.23}&\textBF{58.47} &\textBF{81.73}&\textBF{56.78}& \textBF{82.23}&\textBF{57.62}  & \textBF{82.23} &57.41 &\textBF{72.37}&\textBF{63.61} \\
     \bottomrule
    \end{tabular}
    
    \label{tab:BT_Results}
\end{table}

\vspace{-3pt}
\subsection{Comparison with state-of-the-art methods}
\vspace{-3pt}

To have a comprehensive comparison, we include several related approaches in our experiments: FedAVG \cite{li2019convergence}, MOON \cite{li2021model}, FedProx \cite{li2020federated}, FedFocal \cite{fedfocal}, Fed\clip{} \cite{lu2023fedclip} and a centralized method.  FedAVG fine-tunes and aggregates all the parameters of the \clip{} image and text encoders. MOON extends FedAVG with a contrastive loss between the previous model and the current model. The FedProx approach adds a proximal term to FedAVG that allows having slight differences between clients and the server. FedFocal replaces the standard CE loss with a focal loss for FedAVG. Fed\clip{} adds an adapter to the \clip{} model and only considers the parameters of this adapter during the fine-tuning and aggregation steps. Finally, the centralized method is designed with only one client holding all the training data. For a fair comparison, the same experimental setting described above is used for all tested methods. We evaluate performance using classification accuracy (ACC) as the primary metric. Additionally, due to the class imbalance in the Real dataset, we also consider balanced accuracy (BACC) for a more comprehensive assessment \cite{brodersen2010balanced}. \textit{Additional implementation details and visualization results can be found in the Supplemental materials}.

\begin{figure*}[t!]
    \centering
    \includegraphics[width=\textwidth]{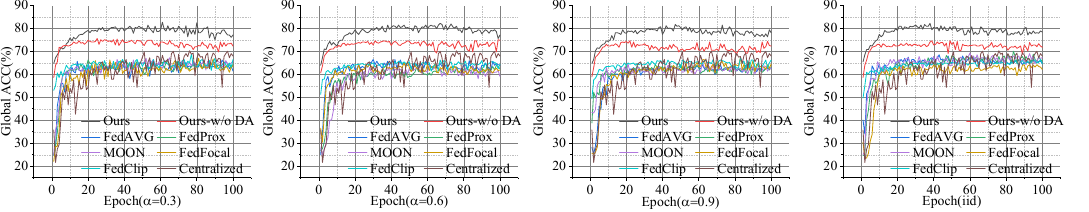}
    \caption{Global testing accuracy (\%) for each round on the BT dataset.}
    \label{fig:BT_training}
\end{figure*}

Table \ref{tab:BT_Results} reports the global classification ACC and BACC for the BT, SC and Real datasets. As can be seen, our FACMIC approach yields a better performance than other FL methods across all datasets, outperforming the second-best federated approach by 15.40\%, 4.03\% and 12.13\% in ACC on BT, SC and Real, respectively. FACMIC also achieves the highest BACC in the Real dataset, outperforming the second-best FL approach by 12.44\%. Figure \ref{fig:BT_training} shows the global test accuracy measured in each communication round. Our method demonstrates a notable increase in accuracy with a minimal number of epochs, highlighting its effectiveness. Specifically, when aggregating parameters in Eq. \ref{Eq:8}, it gives a larger weight to clients with more training samples. In the non-iid setting, this enables the global model to learn from the most knowledgeable clients, preventing performance degradation. The lower performance of other methods in the iid setting could be due to the need of fine-tuning the entire image encoder or to the shallower adaptation module architecture (FedCLIP). Our findings suggest that the \clip{} model can achieve comparable performance in some medical image classification tasks.

\vspace{-3pt}
\subsection{Ablation study}
\vspace{-3pt}

To validate the effectiveness and generalizability of our model, a series of ablation studies were conducted following the same experimental setting. As reported in Table \ref{tab:ablation_DA}, adding the domain adaptation loss $\loss_{\mr{DA}}$ leads to a better classification accuracy in the BT dataset, in all situations. These results demonstrate the need to address the problem of shifting data distribution among clients. 

\begin{table*}[t!] \scriptsize
    \begin{floatrow}
    \capbtabbox{
    \setlength{\tabcolsep}{0.16cm}
    \begin{tabular}{ccccc}
    \toprule
     $ \loss_{\mr{DA}}$  & $\alpha\!=\!0.3$ & $\alpha\!=\!0.6$ &$\alpha\!=\!0.9$ &$iid$   \\
     \midrule
      \checkmark &82.48 & 82.23 &81.73 & 82.23 \\
      \ding{55} &75.63 &75.12   & 74.62 & 74.62   \\
     \bottomrule
    \end{tabular}
    }{
    \caption{Impact of the domain adaptation loss (ACC\%).}
    \label{tab:ablation_DA}
    }
    \capbtabbox{
    \setlength{\tabcolsep}{0.12cm}
    \begin{tabular}{lccccc}
    \toprule
    Batch size&4 &8& 16 & 32 &Avg \\
     \midrule
     BT\,($iid$)&79.69 &80.20 &81.22 & 82.23& 80.84  \\
     Real& 63.39 & 67.32  &69.68 &72.37 & 68.19 \\
     \bottomrule
    \end{tabular}
    }{
 \caption{Impact of batch size (ACC\%).}
 \label{Tab2}
    }
\end{floatrow}
\end{table*}

We also studied the performance of our method for different batch sizes (from 4 to 32), as using large batches is not practical for less capable devices. For sizes of 4 and 8, the adaptation loss is rescaled by a factor of 1/10 since it is too large compared with the contrastive loss in those cases. Table \ref{Tab2} shows the global testing ACC using BT ($iid$) and Real dataset. As one can see, our model is robust to batch size on the BT data, however, using a too small batch size for the large-scale SC data can result in negative adaptation.

To assess our method's generalization ability, we use the fine-tuned global model trained on the BT dataset under iid condition to perform classification directly on another brain tumor dataset (BT2) without any fine-tuning. The BT2 dataset comprises 7023 samples obtained from figshare, SARTAJ, and Br35H \cite{BrainTumor_MRI_2, Cheng2017,sartaj,BrainTumor_MRI_3}, and has the same classes as the BT dataset. We tested these methods using the whole dataset. As reported in Table \ref{tab:generlization}, our method achieves the highest generalization accuracy of 94.42\% on BT2.

\begin{table}[t!] \scriptsize
    \centering
    \caption{Global testing accuracy (\%) using BT2 dataset.}
    \setlength{\tabcolsep}{0.2cm}
    \begin{tabular}{ccccccc}
    \toprule
    FedAVG & FedProx & MOON & FedFocal & Centralized & Fed\clip{} &Ours \\
     \midrule
      84.18 &84.72 &82.32 &79.99 &86.12 & 91.7  & \textBF{94.42}   \\
     \bottomrule
    \end{tabular}
    \label{tab:generlization}
\end{table}


\vspace{-3pt}
\section{Conclusion}\label{S:5}
\vspace{-3pt}

We explored the usefulness of VLMs for medical imaging in FL and presented a novel FACMIC framework that combines a feature attention module to reduce communication costs and a domain adaptation strategy to minimize data distribution differences between each client. Experimental results in brain tumor and skin cancer classification tasks demonstrate the superior performance of FACMIC compared to state-of-the-art FL approaches. 

\begin{credits}
\subsubsection{\ackname}
This research was funded by the National NSFC (82260360), the Guilin (20222C264164), and the Guangxi talent (2022AC18004, 2022AC21040). 

\subsubsection{\discintname}
The authors have no competing interests to declare that are relevant to the content of this article. 
\end{credits}

\bibliographystyle{unsrt}
\bibliography{Paper-1577}

\end{document}